\documentclass[10pt,twocolumn,letterpaper]{article}

\usepackage{iccv}
\usepackage{times}
\usepackage{epsfig}
\usepackage{graphicx}
\usepackage{subfigure}
\usepackage{amsmath}
\usepackage{amssymb}
\usepackage{multirow}
\usepackage[ruled,vlined]{algorithm2e}
\SetAlFnt{\footnotesize}

\usepackage[breaklinks=true,bookmarks=false]{hyperref}

\iccvfinalcopy % *** Uncomment this line for the final submission

\def\iccvPaperID{7} % *** Enter the ICCV Paper ID here

\ificcvfinal\fi

\begin{document}

%%%%%%%%% TITLE
\title{A System-Level Solution for Low-Power Object Detection}

\author{Fanrong Li\textsuperscript{1,}\textsuperscript{2}\thanks{These authors contributed equally.}  , Zitao Mo\textsuperscript{1,}\textsuperscript{2$*$}, Peisong Wang\textsuperscript{1}, Zejian Liu\textsuperscript{1,}\textsuperscript{2$*$},\\
Jiayun Zhang\textsuperscript{1$*$}, Gang Li\textsuperscript{1,}\textsuperscript{2}, Qinghao Hu\textsuperscript{1}, Xiangyu He\textsuperscript{1,}\textsuperscript{2}, Cong Leng\textsuperscript{1,}\textsuperscript{3},\\
 Yang Zhang\textsuperscript{1,}\textsuperscript{3}, Jian Cheng\textsuperscript{1,}\textsuperscript{2,}\textsuperscript{3,}\textsuperscript{4}\\
\textsuperscript{1}Institute of Automation, Chinese Academy of Sciences\\
\textsuperscript{2}University of Chinese Academy of Sciences, \textsuperscript{3}AiRiA\\
%textsuperscript{3}AiRiA\\
\textsuperscript{4}CAS Center for Excellence in Brain Science and Intelligence Technology\\
{\tt\small \{lifanrong2017, mozitao2017\}@ia.ac.cn, \{gang.li, jcheng\}@nlpr.ia.ac.cn}}
%\and
%Yang Zhang\\
%AIRIA\\

\maketitle
\ificcvfinal\thispagestyle{empty}\fi

%%%%%%%%% ABSTRACT
\begin{abstract}
   Object detection has made impressive progress in recent years with the help of deep learning. However, state-of-the-art algorithms are both computation and memory intensive. Though many lightweight networks are developed for a trade-off between accuracy and efficiency, it is still a challenge to make it practical on an embedded device. In this paper, we present a system-level solution for efficient object detection on a heterogeneous embedded device. The detection network is quantized to low bits and allows efficient implementation with shift operators. In order to make the most of the benefits of low-bit quantization, we design a dedicated accelerator with programmable logic. Inside the accelerator, a hybrid dataflow is exploited according to the heterogeneous property of different convolutional layers. We adopt a straightforward but resource-friendly column-prior tiling strategy to map the computation-intensive convolutional layers to the accelerator that can support arbitrary feature size. Other operations can be performed on the low-power CPU cores, and the entire system is executed in a pipelined manner.
   As a case study, we evaluate our object detection system on a real-world surveillance video with input size of 512$\times$512, and it turns out that the system can achieve an inference speed of 18 fps at the cost of 6.9W (with display) with an mAP of 66.4 verified on the PASCAL VOC 2012 dataset.
\end{abstract}

%%%%%%%%% BODY TEXT
\section{Introduction}

Since AlexNet \cite{Krizhevsky2017ImageNet} won the 2012 large-scale image recognition contest, Deep Convolutional Neural Networks (DCNNs) have shown increasing performance in various computer vision tasks.
CNN's impressive performance is mainly due to its high complexity and capacity, in other words, the great number of parameters and computations. 
Therefore, high-performance hardwares such as GPUs (clusters) are often utilized for acceleration.
However, as for embedded and mobile devices such as drones, security cameras, and smart glasses, GPU-based solutions are not the best choice due to the limitation of volume and power consumption. 
In addition, modern GPUs that designed for general propose processing are not flexible enough to deal with low-bit integer values less than 8-bit without efforts on tuning the codes.
As a result, FPGA-based accelerators are gaining popularity in recent years for both industrial and academic communities.

%The performance of CNNs depends heavily on the size (or complexity) of the network. 
%In general, deeper networks with complex connections tend to work better on accuracy. 
%In the field of DNN accelerator, some existing work selects a larger network \cite{GoingDeeper, Suda:2016:TOF:2847263.2847276}, but this can lead to significant computation and storage overhead, which may not be a practical choice.
%When deploying on FPGA, the massive multiply-add operations involved in the network require a great amount of on-chip DSP blocks which are often limited on most off-the-shelf FPGAs \cite{Zhang:2015:OFA:2684746.2689060}. Besides, the huge volume of intermediate data of large network imposes big pressures on on-chip storage, thus off-chip data transfers should be executed frequently, which is inefficient in terms of power consumption and latency.
%When considering real-time processing, an oversized CNN model is also unhelpful for speed.
%Some designs proposed to use a smaller network \cite{Zhao:2017:ABC:3020078.3021741, Umuroglu:2017:FFF:3020078.3021744}, but at the expense of accuracy degradation.
%In addition, most existing accelerators are difficult to achieve high performance under limited resource budgets, especially in low-power settings.

As for memory efficiency, we find that the advantages of the recent depthwise convolution \cite{XceptionDL, MobileNet} are apparent. Unlike traditional convolution, in depthwise convolution, each output feature map relies solely on a single input feature map in the previous layer, which dramatically reduces the amount of computations and the demand of on-chip storage. In terms of resource and energy efficiency, recent logarithmic computation \cite{ShiftCNN, IncrementalNQ, Lee2017LogNetEN} has shown its promise. It quantizes the weight as power-of-two in order to efficiently translate multiplication into bit shift operation, which can get rid of the limitation of insufficient on-chip DSP blocks.

Considering the advantages of depthwise convolution and logarithmic computation mentioned above, we put forward an end-to-end hardware-software co-design for low-power object detection on resource-constraint FPGA. Our proposed solution can achieve relatively high performance under extremely low resource budget while retaining considerable accuracy.
The contribution of this work can be summarized as follows:
\begin{itemize}
	\item We propose a dedicated object detection accelerator for customized MobileNet-SSD \cite{Liu2016SSD,MobileNet} algorithm through software-hardware co-design. Specifically, we quantize the activations and weights to 4-bit integer and 3-bit power-of-two integer respectively, and present a fused-layer architecture with shift-based processing elements.
	\item We adopt a column-prior strategy to map the detection network to the accelerator, which can reduce resource consumption. Besides, a hybrid dataflow is introduced to reuse output or weights according to the heterogeneous property of different layers.
	\item We highlight the entire pipeline of our heterogeneous system design, including hardware accelerator, host processing and thread management of the main processor, and describe each stage in details.
	\item We verify the performance of our design on heterogeneous devices Ultra96 SoC that targets to IoT applications. Experiments show that the entire system can reach an inference speed of 18 fps at the cost of around 6.9W.
\end{itemize}

The rest of the paper is organized as follows. Section \ref{sec2} describes the quantization algorithm, with which we quantize weights to the power-of-two and enables resource-friendly shift-base multiplications. Section \ref{sec3} briefly presents the overall system architecture. Section \ref{sec4} introduces the architecture of the dedicated accelerator, including Processing Elements (PEs), tiling strategy, and dataflow. Section \ref{sec5} reports the experimental results as well as multithread management on low-power CPUs.
%%%%%%%%% BODY TEXT
%-------------------------------------------------------------------------
\section{Quantization} \label{sec2}
\begin{figure}[th]
	%\begin{center}
	\centering
	\includegraphics[width=0.35\columnwidth]{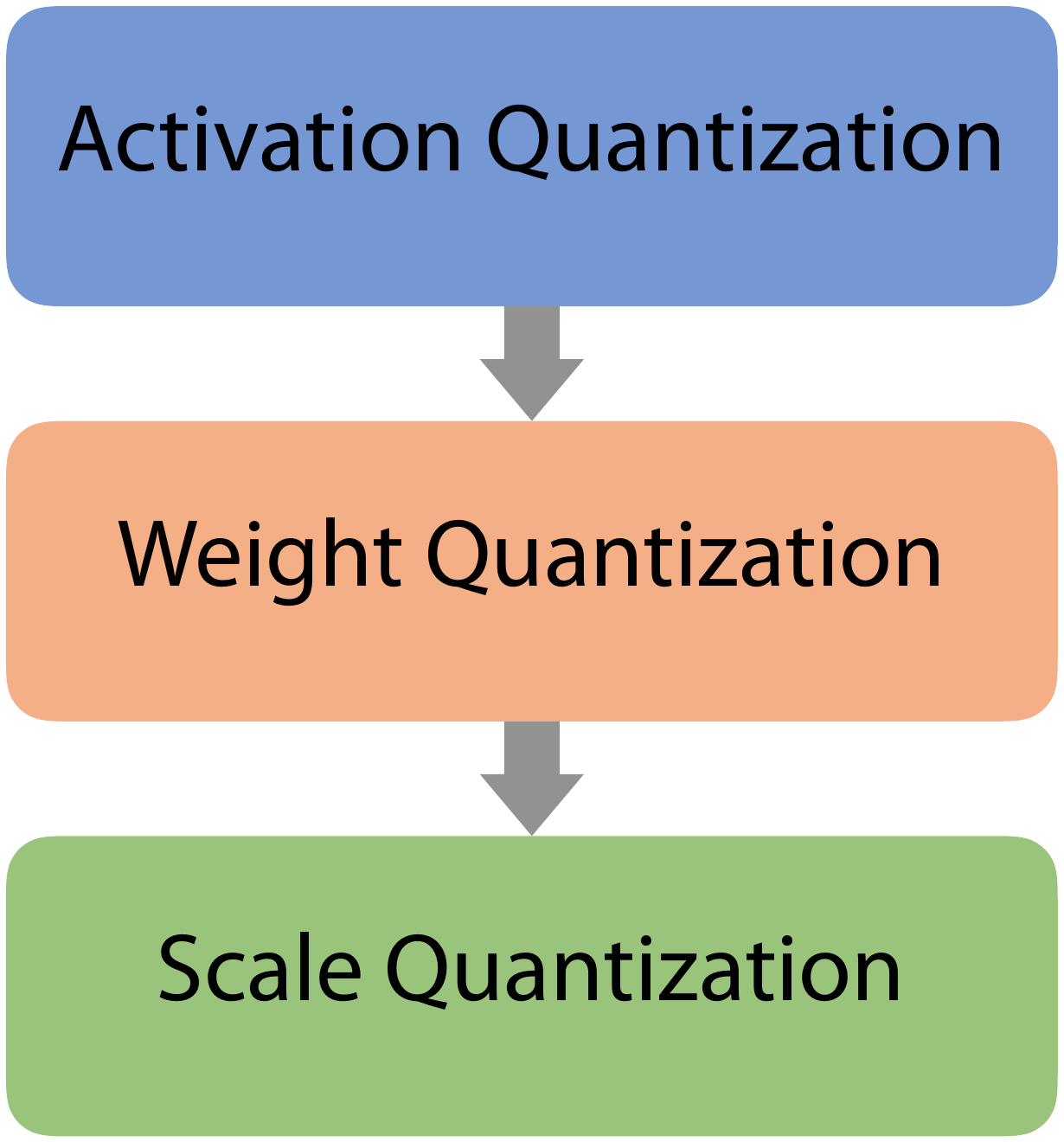}
	%\end{center}
	\caption{Three-Step Quantization Pipeline.}
	\label{fig:ttq}
\end{figure}

To make the CNN model compatible with our hardware architecture design, we introduce a three-step quantization method, i.e., uniform activation quantization, power-of-two weight quantization as well as scale quantization, as illustrated in Figure \ref{fig:ttq}.
It is worth noting that through the proposed three-step quantization, 
%\textit{all computing can be transformed into fixed-point operations, without any floating-point values}
\emph{all computing can be transformed into fixed-point operations, without any floating-point values}.
%We will show that these quantization
%problems can be solved by the proposed Iterative Quantizer in section \ref{sc:floyd}.

\subsection{Uniform Activation Quantization}
\label{sc:activation_quantization}
%For a given full-precision neural network with ReLU (Rectified Linear Units) activation function,
%the activations are all positive.
For $M$-bit activation quantization, we want to
quantize all the positive activations into the set $\mathcal{A}=\{0, 1, 2, \cdots, 2^M-1\}$.
As with many other fixed-point quantization methods, we also introduce a scaling factor $\alpha$ to lower the quantization error, making the quantization set into
\begin{equation*}
	\mathcal{A}=\{0, 1, 2, \cdots, 2^M-1\}*\alpha
\end{equation*}

To turn all activations into fixed-point numbers, we can quantize the floating-point activation to the nearest point in the set $\mathcal{A}$. The $2^M-1$ quantization thresholds can be set to the medians of two successive quantized values:
\begin{equation}
	\begin{aligned}
		& t_i=\frac{(i-1)\alpha+i\alpha}{2}=(i-\frac{1}{2})\alpha\\
		& \text{for } i=1,\cdots, 2^M-1
	\end{aligned}
\end{equation}

%$\mathcal{A}=\{0, 1, 2, \cdots, 2^M-1\}*\alpha$.
Thus the quantization function $Q_a$ can be formulated as
\begin{equation}
	\label{eq:activation_quantization}
	Q_{A}(x)=\left\{
	\begin{array}{cll}
		{(2^M-1)\alpha}  & x>t_{2^M-1}, \\
		{i\alpha} & x\in(t_{i}, t_{i+1}], \\
		{0}   & x\le{t_1}
	\end{array}
	\right.
\end{equation}

%The difficulty of the quantization problem is to find the scaling factor $\alpha$.
%We will see how the optimal $\alpha$ can be determined using Iterative Quantizer
%in section~\ref{sc:floyd}.

\subsection{Power-of-Two Weight Quantization}
%In this subsection, we will discuss how to turn all the full-precision weights of a pre-trained
%network into pwoer-of-two values. For the convolutional neural networks considered in this paper,
%the weights for a convolutional layer is a 4-D tensor, i.e., $\mathcal{W}\in\mathbb{R}^{w\times h\times c\times n}$.
%Here, $w$ and $h$ represent the kernel width and height, $c$ and $n$ represent the numbers of input and output channels.
%For simplicity, we do not differentiate the dimensions and treate the whole weights as a vector.

For weight, we utilize power-of-two quantization. In this way, the floating-point multiplications within the convolution can be transformed into shifting operations, which can dramatically lower the complexity of CNN and hardware design. The 4-D weight tensor consists of $n$ kernels of size $w\times h\times c$, which are quantized by using different scaling factors.
More specifically, the 4-D tensor $\mathcal{W}\in\mathbb{R}^{w\times h\times c\times n}$ is reshaped into a matrix $\mathbf{W}\in\mathbb{R}^{(w*h*c)\times n}$, where each column $\mathbf{w}_i\in\mathbb{R}^{w*h*c}$ corresponds to a 3-D kernel. To lower the quantization error, a floating-point scaling factor $\beta_i$ is introduced for each kernel $\mathbf{w}_i$, i.e., for $N$-bit quantization, the problem is to select weight values from the set
\begin{equation*}
	\mathcal{B}_i=\{0, \pm 2^0, \pm 2^1, \cdots, \pm 2^{2^{N-1}-2}\}*\beta_i
\end{equation*}

Here we also use the nearest quantization and the $2^N-2$ quantization thresholds can also be determined by the medians of two successive quantized values, as in the activation quantization.
%The quantization thresholds as well as the quantization function are determined by the only parameter
%$\beta_i$, which can also be solved by the Iterative Quantization algorithm discribed in section~\ref{sc:floyd}.

\subsection{Scale Quantization}
By activation and weight quantization, the convolution can be performed with only fixed-point operations. However, the whole network still requires floating-point operations due to the introduced scaling factors, bias term of convolution, as well as some other layers like Batch Normalization.
%These time and resource consuming floating-point operations
%could become the bottleneck of the whole network, especially in our quantization scheme, where all multiplications
%are transformed into lightweight bit-shifting operations.
To further eliminate the above mentioned floating-point operations, we introduce the scale quantization, which consists of two parts:

\begin{table}
	\small
	\caption{Quantization results on ImageNet classification (top-1 accuracy).
		The \#Act., \#Wei. and \#Sca. represent the number of bits for activations, weights, and scaling factors, respectively.}
	\vspace{1em}
	\centering
	\begin{tabular}{|c|c|c|c|c|}
		\hline
		Model  & \#Act. & \#Wei. & \#Sca. & Accuracy \\
		\hline
		\hline
		MobileNet & Full & Full & Full & 70.1 \\
		\hline
		MobileNet & 8 & 3 & 8 & 68.3\\
		\hline
		MobileNet & 4 & 3 & 8 & 68.1\\
		\hline
	\end{tabular}
	\label{q_result}
\end{table}

\textbf{Scale merge:}
%In a typical convolutional neural netowk, the floating-points including the scaling factor as well as
%other floating-points like the parameters of the Batch Normalization layers
%can be merged together. We will show how this
%can be done in this section.
For the $l$-th layer, the input activation $X$ can be represented by $X=\alpha \hat{X}$, where $\hat{X}$ is the fixed-point version of $X$ and $\alpha$ is the scaling factor.
Similarly, the $\mathbf{w}=\beta\hat{\mathbf{w}}$ where $\hat{\mathbf{w}}$ is one of the fixed-point kernels.
For simplicity, we discard the kernel index.
%By taking the Batch Normalization term into consideration, 
Considering the Batch Normalization term, the convolutional layer can be represented by the following equation:
\begin{equation}
	\label{forward}
	\begin{aligned}
		Y=\alpha'\hat{Y}&=Q_A(BN(\beta\hat{\mathbf{w}}\otimes\alpha\hat{X}))\\
		&=Q_A(\gamma\alpha\beta\hat{\mathbf{w}}\otimes\hat{X}+b)\\
		&=Q_A(a\hat{\mathbf{w}}\otimes\hat{X}+b)
	\end{aligned}
\end{equation}
where $Y$ is the output activation, $\hat{Y}$ is the fixed-point version of output activations, and the $\alpha'$ is
the scaling factor for outputs. $BN(x)=\gamma x + b$ is the batch normalization layer and $\otimes$
is the convolution.

To further merge out the output scaling factor, we can divide both sides of Eq. \ref{forward} by $\alpha'$, resulting in the following equation:
\begin{equation}
	\label{forward2}
	\begin{aligned}
		\hat{Y}&=\hat{Q}_A(\frac{a}{\alpha'}\hat{\mathbf{w}}\otimes\hat{X}+\frac{b}{\alpha'}))\\
		&=\hat{Q}_A(a'\hat{\mathbf{w}}\otimes\hat{X}+b')
	\end{aligned}
\end{equation}

Note that in the activation quantization function need to be changed accordingly.
By defining $\hat{t}_i=\frac{t_i}{\alpha'}$, the new quantization function becomes:
\begin{equation}
	\label{eq:new_activation_quantization}
	\hat{Q}_{A}(x)=round(clip(x, 0, 2^M-1)),
\end{equation}
where $round(x)$ is the rounding operation, and $clip(x,u,v)$ clips $x$ within $u$ and $v$.

\textbf{Scale quantization:} In Eq. \ref{forward2}, only the $a'$ and $b'$ are floating-points.
Note that Eq. \ref{forward2} only coresponds to one 3-D kernel, for the convolutional layer, there are $n$ pairs of $a'$ and $b'$, denoted by $\mathbf{a'}$ and $\mathbf{b'}$.
In the scale quantization, we need to quantize these values into fixed-point numbers.

During the scale quantization, no scaling factors could be incorporated. 
However, direct quantizing of $\mathbf{a'}$ and $\mathbf{b'}$ will introduce large quantization error. Here we search for the binary point position, resulting in the following set to be quantized into:
\begin{equation*}
	\mathcal{C}=\{0, \pm 1, \pm 2, \cdots, \pm 2^{K-1}-1\}*2^{d},
\end{equation*}
where $d$ represents the binary point position.
More specifically, when $d$ go throught from 0 to -15, we find the best $d$ that minimize the
quantization error for $\mathbf{a'}$ and $\mathbf{b'}$.

\begin{figure}[t]
	\begin{center}
		\includegraphics[width=0.75\columnwidth]{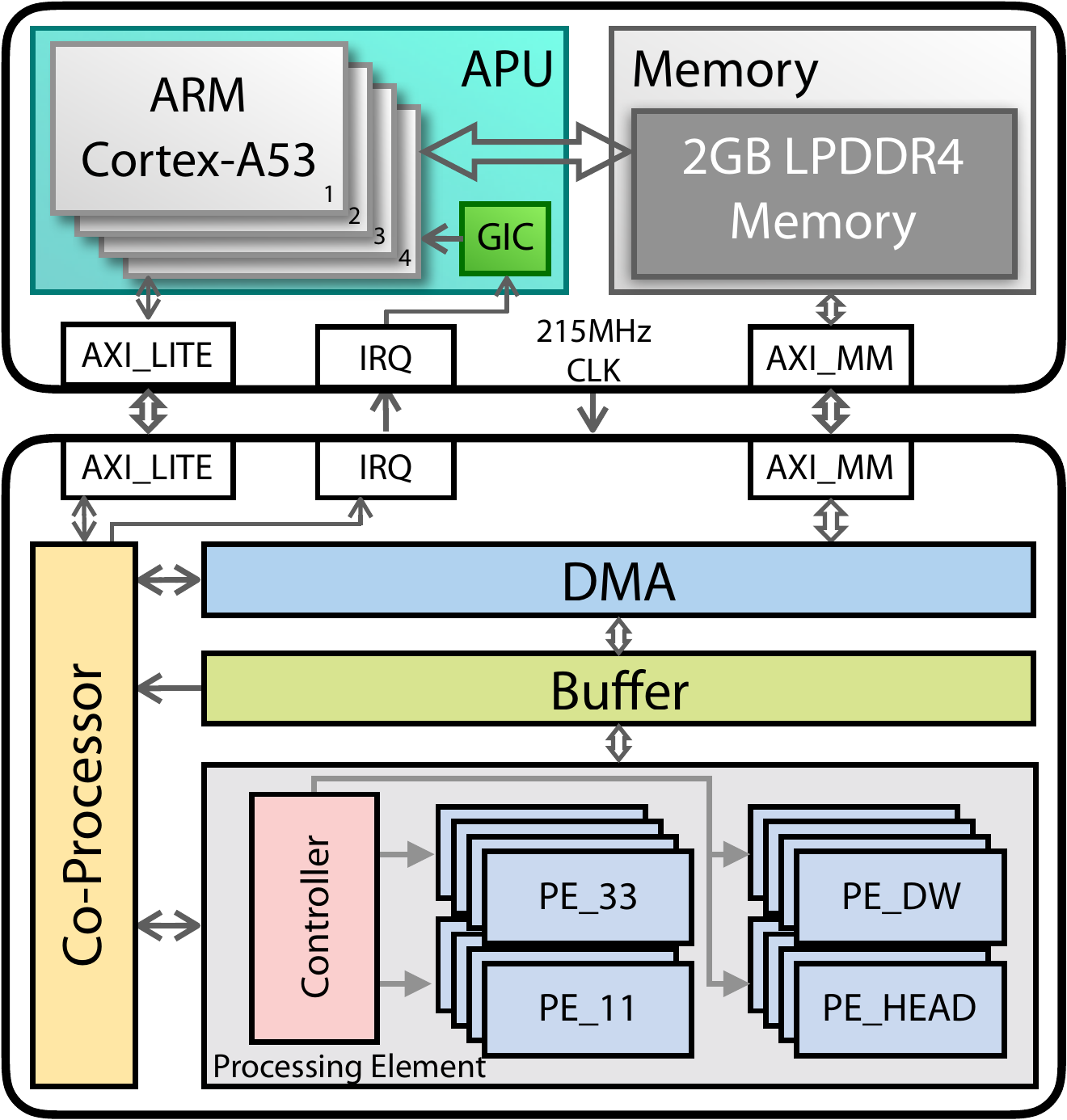}
	\end{center}
	\caption{Architecture of the entire system.}
	\label{System}
\end{figure}

\subsection{Optimization}
%In this section, we describe how to find the optimal scaling factors for the activation/weight quantization.
%Given a set of floating-point data points (from the activations or weights of the pre-trained networks),
%our objective is to find the optimal scaling factor which minimize the quantization error of these data points
%and the quantized counterparts.

The optimization problem can be solved efficiently using Lloyd's algorithm.
%It is similar to the k-means algorithm, which iteratively conduct data points
%assignment and centroids update.
Take the activation quantization problem of section~\ref{sc:activation_quantization} for example, during the assignment step, all activation data points are quantized into the nearest fixed-point values in the set of $\mathcal{A}$ according to the quantization function $Q_A(x)$.
In the update step, the new scaling factor can be obtained by solving a one-dimensional optimization problem:
\begin{equation}
	\alpha^*=\underset{\alpha}{\text{argmin}}\sum_x{(x-Q_A(x))^2}
\end{equation}
By iterative quantization, we could find the optimal scaling factors as well as the quantized values.

\label{sc:floyd}

After the activation quantization and weight quantization, we need to fine-tune the whole network to restore accuracy.

%However, due to the fixed-point quantization, the network becomes non-differentiable.
%Like many fixed-point quantization methods,
%we use the Stright-Through approximator \cite{bengio2013estimating} for the gradient approximation. For the
%weight update, we keep two versions of weights, one is the fixed-point version of weights, which
%are used for the forward-backward computation, and another floating-point version of weights,
%which are used for the gradient accumulation.
%More details can be found in the work of
% \cite{courbariaux2015binaryconnect, hubara2016binarized}.

\subsection{Performance}
%The quantization results of the proposed three-step quantization approach are given in this section.
The experiments are conducted on the ImageNet classification benchmark, results are shown in Table~\ref{q_result}. The results illustrate that the three-step quantization approach has only minimal accuracy drop compared with the floating-point counterpart.

\section{System Architecture} \label{sec3}

Our detection network targets to run on the Ultra96 development board, 
%on a low-power embedded device such as Ultra96 board, 
which is a heterogeneous embedded system containing both programmable logic and low-power CPU cores. 
%There is an UltraScale+ MPSoC with 4 Cortex A53 CPUs. 
A 2GB DDR4 is shared by Programmable Logic (PL) and Processing System (PS). 
%with the support of the AXI protocol. 
Since convolutional layers dominate most of the inference time, we implement a dedicated CNN accelerator with the Programmable Logic. 

The entire system includes the following functional layers. 
Data forward layer: decode video streams. 
Encode layer: organize data into the specific pattern for FPGA accelerator. %transfer data to a proper layout for FPGA accelerator. 
FPGA layer: perform all convolution on the dedicated accelerator. 
Decode layer: organize extracted features from the accelerator to the storage pattern for CPU. %transform extracted features from the accelerator to a proper layout for CPU. 
Mbox-conf-reshape layer: reshape bounding boxes. 
Mbox-conf-softmax layer: softmax layers of the detection. 
Mbox-conf-flatten layer: reshape data. 
Detection and visualize layer: generate detecting results and display on the screen. 
%Besides the FPGA layer, all other layers are completed on the CPU. 
All the layers except for FPGA layer are executed on CPU. 
All operations before the FPGA layers are referred to as pre-processing, while those operations after the FPGA layer are post-processing.

At the very beginning, images together with the weights and instructions of a specific CNN are stored in DDR. The CPU initiates a calculation request and transfer instructions to the accelerator through AXI. The accelerator receives instructions and completes all convolution computation. Note that the accelerator has its own instruction set, 
%once it receives initial instructions 
and it can complete the calculations independently unless interrupted by exceptions. %Extracted features 
Results of the FPGA layer are sent back to CPU for post-processing. Multi-thread technique is exploited to make the most use of 4 low-power ARM cores. The entire system works in a pipelined manner, and the system architecture is shown in Figure \ref{System}.

\section{Dedicated Accelerator} \label{sec4}
In this section,  we first describe the overall architecture of our accelerator, which exploits multiple PEs for high computing parallelism. Then the design of PE is introduced.  After that, the column-prior tiling strategy is presented to support the arbitrary size of input feature maps under limited resources. Finally, a hybrid dataflow is proposed for more efficiency.

\subsection{Overall Architecture}

\begin{figure}[t]
	\begin{center}
		\includegraphics[width=0.9\columnwidth]{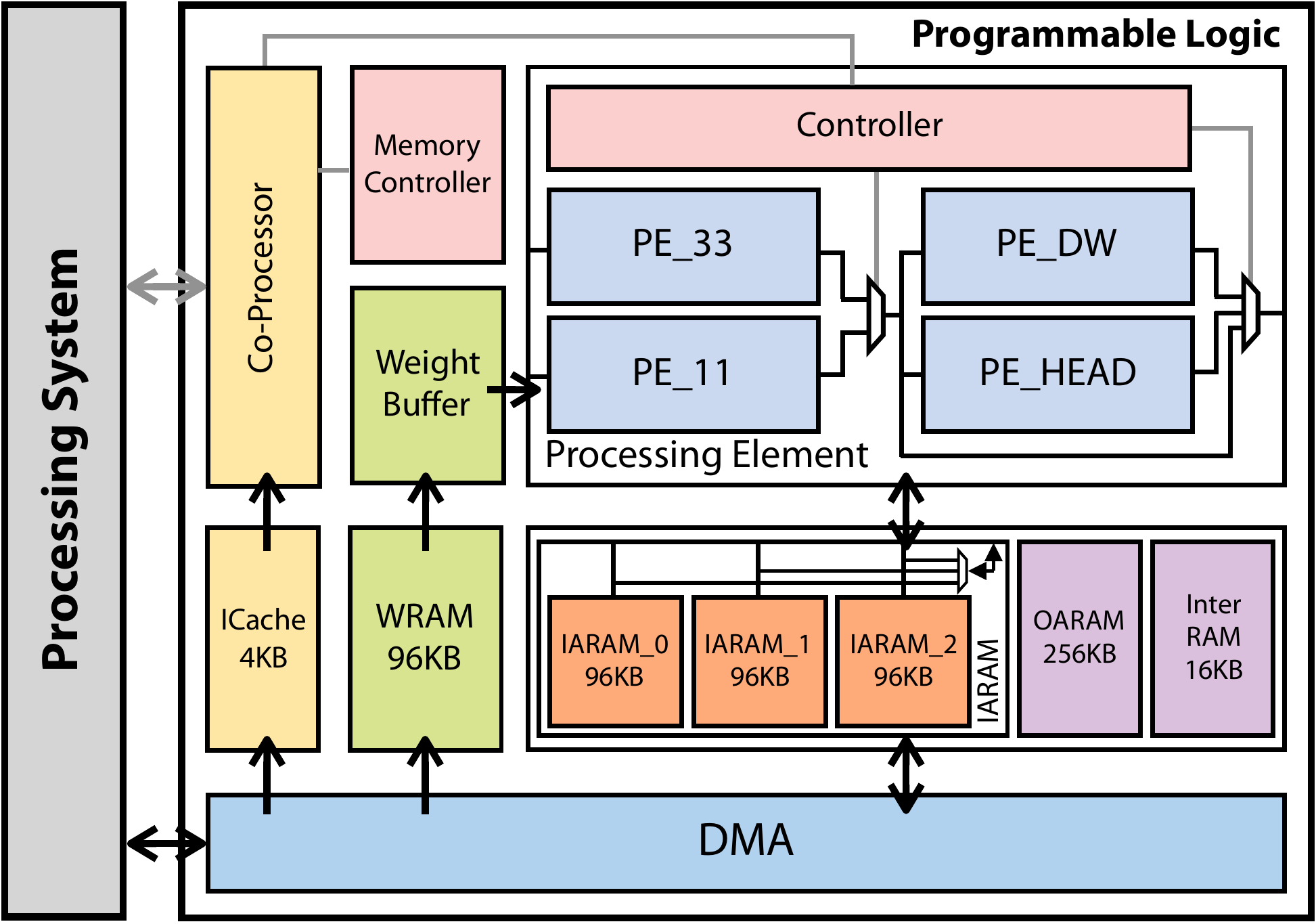}
	\end{center}
	\caption{Architecture of dedicated CNN accelerator with only one for each type of PEs.}
	\label{Architecture}
\end{figure}

Figure \ref{Architecture} shows the overall architecture of our accelerator with different types of PEs inside.
The Co-Processor module controls the entire computation flow. It parses instructions to generate control information for the Memory Controller and different kinds of PEs. 
The addresses of activations and weights are calculated by the Memory Controller, with which all kinds of data can be sent to the proper destinations. 
Prefetching is enabled since we implement a 4KB instructions cache inside the Co-processor. 
Note that some cache features are unavailable in this design because they are unnecessary for a specific accelerator without branch and jump instructions. 
Controllers for different types of PEs generate control signals according to the control information received from Co-Processor. 
IARAM and OARAM  are used to store the intermediate feature maps during computation, where IARAM is implemented with three banks, providing sufficient bandwidth to complete the $3\times 3$ convolution more efficiently. And the IARAMs and OARAM can be logically swapped between the computation of two adjacent layers.
We implement two-level weight caches (Weight buffer and WRAM) with on-chip registers and BRAMs, which can provide sufficient bandwidth for computing. 

\subsection{Processing Elements}

Heterogeneous nature of 1$\times$1 convolution and depthwise convolution may make the reuse of processing elements costly, so reusing PEs does not necessarily lead to benefits and is contrary to our original intention to design a dedicated low-power accelerator. 
Therefore, PEs are specialized for different kinds of convolutional layers, i.e., 3$\times$3 convolution (PE\_33), 1$\times$1 convolution (PE\_11), and depthwise convolution (PE\_DW) for the consideration of reducing the control complexity and improving hardware efficiency. 
%To complete the location offset computing for object detection, extra PE HEAD is necessary.
To efficiently compute the location offsets in the detection algorithm, PE\_HEAD is necessary.
Each type of PEs is mainly composed of multipliers and reduction trees, %as well as modules of the ReLU and Batch Normalization, which can be bypassed. 
as well as modules that can selectively execute the ReLU and  Batch Normalization functions.
Each PE processes with only one kernel at a time.

\begin{figure}[t]
	\centering
	\subfigure{
		\begin{minipage}{0.9\linewidth}
			\includegraphics[width=7.cm]{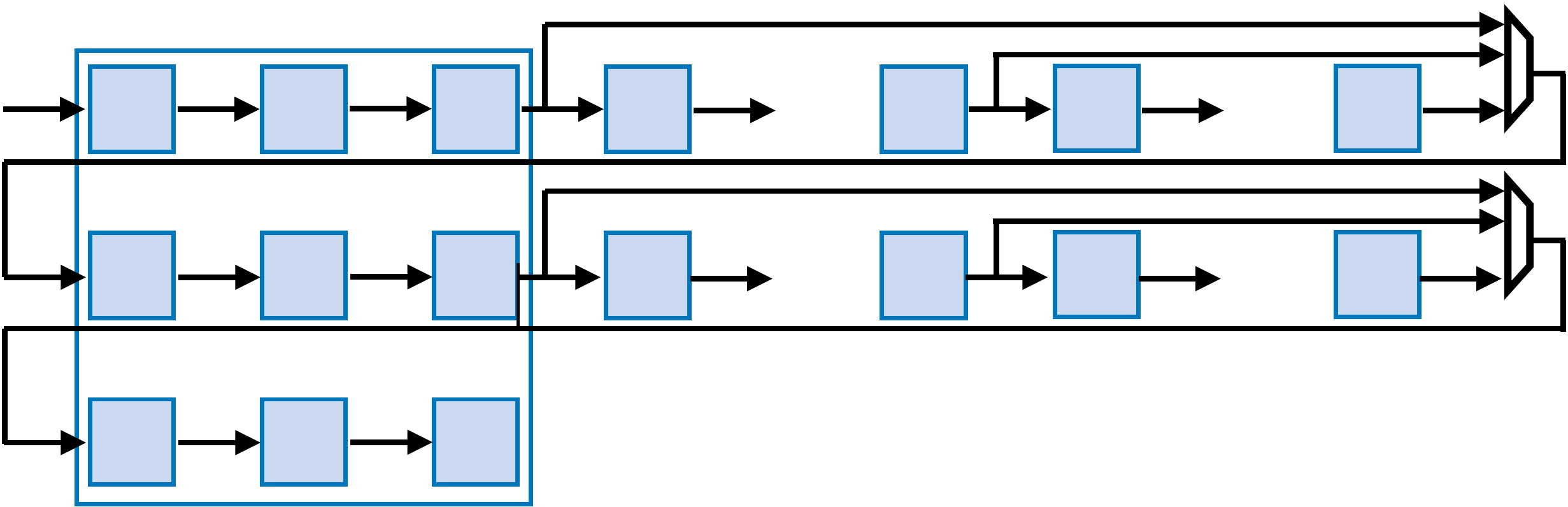}
			\centerline{\small{(a) Line buffer convolution.}}
		\end{minipage}
	}
	
	\subfigure{
		\begin{minipage}{0.35\linewidth}
			\centerline{\includegraphics[width=2.4cm]{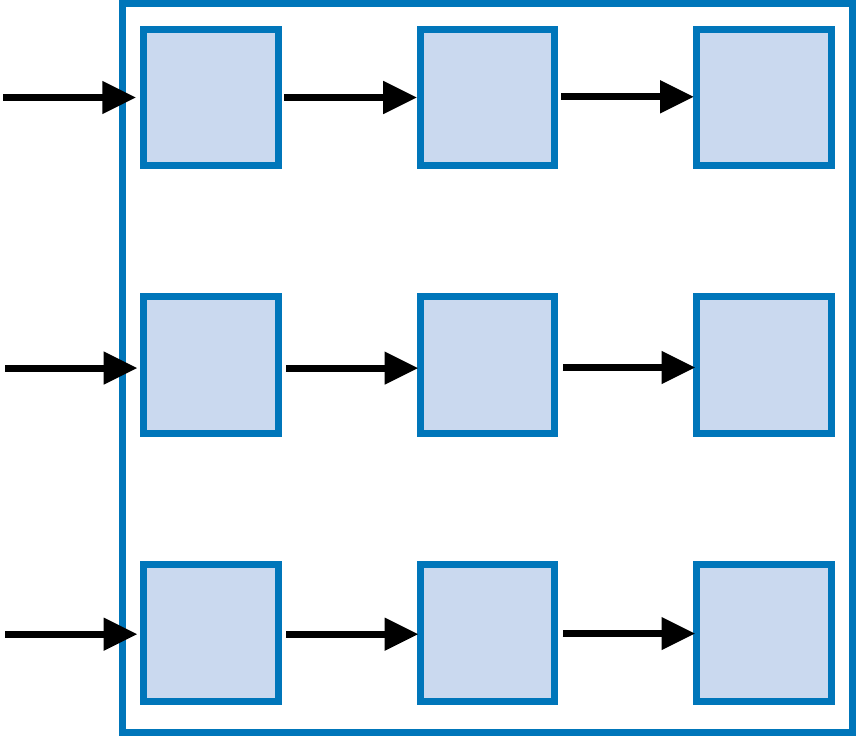}}
			% \vspace{0.2em}
			\centerline{ \small{(b) PE\_33, stride=1.}}
		\end{minipage}
		\begin{minipage}{0.6\linewidth}
			\centerline{\includegraphics[width=2.47cm]{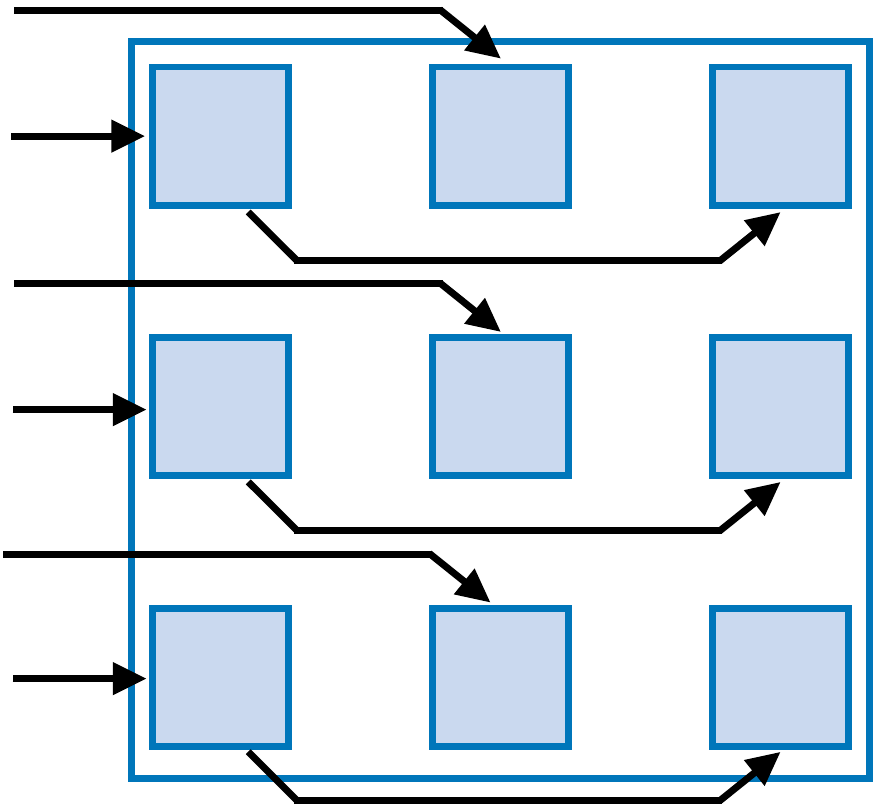}}
			\centerline{ \small{(c) PE\_33, stride=2.}}
			\label{conv3_s2}
		\end{minipage}
	}
	\vspace{0.8em}
	\caption{The implementation for 3$\times$3 convolution with different strategies: (a) Line buffer convolution; (b) Our implementation of 3$\times$3 convolution with stride=1; (c)  Our implementation of 3$\times$3 convolution with stride=2}
	\label{PE}
\end{figure}

Different from some previous work using line buffer, we implement 3$\times$3 convolution in PE\_33 more efficiently, as shown in Figure \ref{PE}. 
The input image is divided into three parts according to row number and stored in three IARAMs. 
During the computation, inputs in three continuous rows can be fetched from different IARAMs simultaneously.
Compared to line buffer implementation, it reduces data-preparing time and register consumption. 
Besides, as for the 3$\times$3 convolution with stride=2, each IARAM can provide higher bandwidth to support jump connection for the registers, as shown in Figure \ref{conv3_s2}.  Therefore, only the necessary calculations are performed, 
%, boosting performance.
which can achieve 4$\times$ speedup than the original convolution based on classic line buffer.

%In addition, depth-width convolution can be fused with others between layers, that allows starting processing as soon as the output of the previous layer, minimizing the memory accesses of feature maps. 

Depthwise convolutional layer can be fused with its adjacent layers in a pipelined manner to speedup computation due to its less data-dependent property. With this insight, in this work, we introduce two types of cascaded PEs to the architecture of our accelerator, which can be summarized as follows.

\begin{itemize}
	\item {\bf PE\_33, PE\_DW.} The results of 3$\times$3 convolution can be sent to PE\_DW directly. Different from PE\_33, PE\_DW are processing with line buffer to accommodate the continuous inflow of data. This manner works in conjunctions with our column-prior tiling strategy to reduce the consumption of registers, which we will present in section \ref{sec:tiling}.
	
	\item {\bf PE\_11, PE\_DW.} Similarly, 1$\times$1 convolution and depthwise convolution can also be processed in a fused manner. During computation, input activations are fetched from one of three input buffers, and the results of 1$\times$1 convolution are sent  to PE\_DW immediately and processed on the fly. The final results are written back to the corresponding output buffer. 
\end{itemize}

\begin{figure}[t]
	\begin{center}
		\includegraphics[width=0.85\columnwidth]{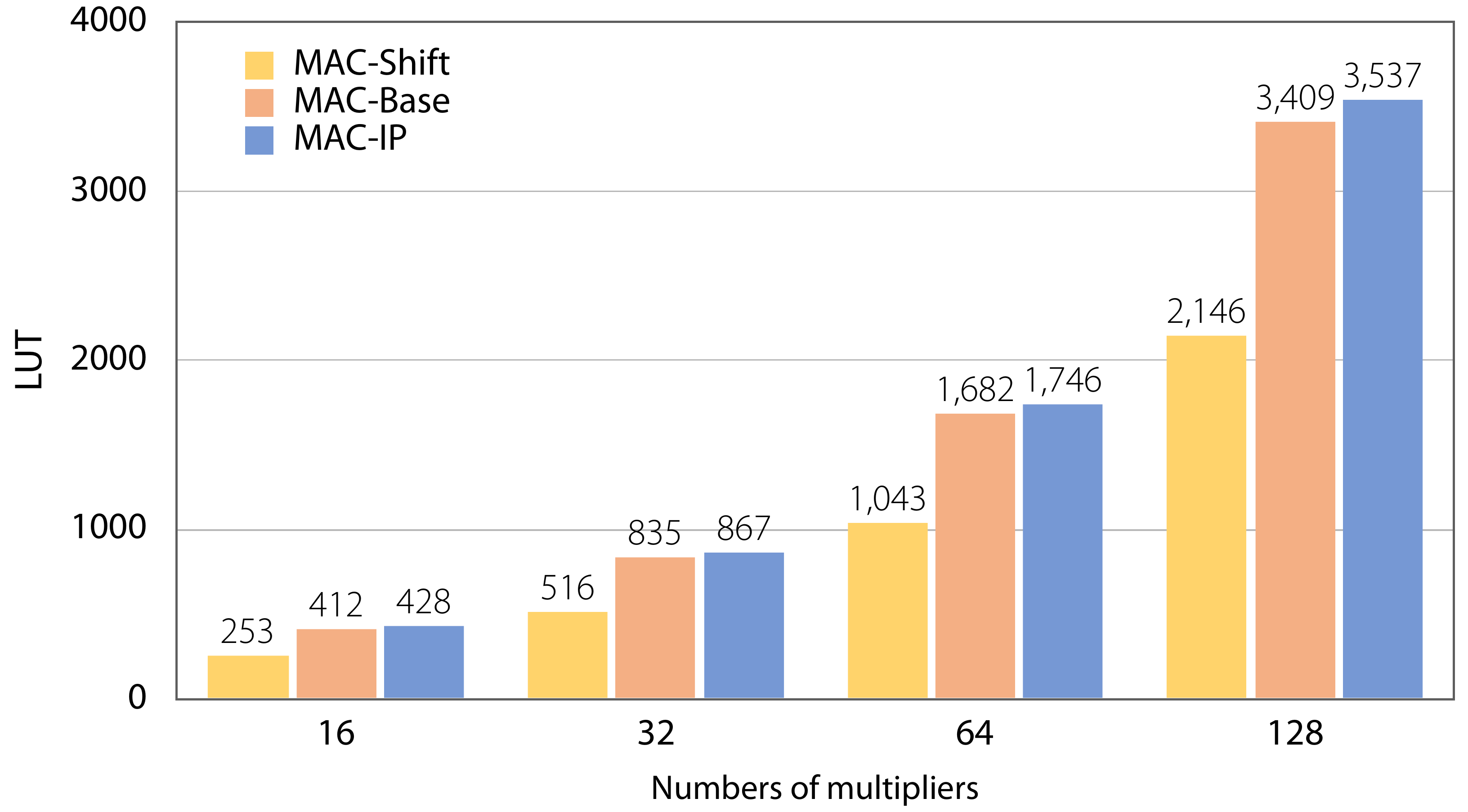}
	\end{center}
	\vspace{-0.3em}
	\caption{LUT consumption of different implementations of MACs. MAC-Shift (activation 4b/weight 3b) is our implementation of multipliers using shift operations, while MAC-Base (4b/4b) is direct multiplication and MAC-IP (4b/4b) is multiplications using Xilinx IP. Reduction trees are also included in all three cases. Note that if we use multipliers, we have to use 4b/4b inputs in order to represent numbers from -4 to 4.}
	\label{Mult}
\end{figure}

As mentioned in section \ref{sec2}, activations and weights of the network are quantized to low bits. Specifically, the weights are quantized to power-of-two, which enables us to replace multipliers with shift operators. Compared with normal multiplications, it can reduce resource and power consumption. We conduct an experiment to verify the benefits of this shift-based multipliers, which shows that shift-based multiplication can reduce the usage of LUT by approximately 40\%, as shown in Figure \ref{Mult}. 

\subsection{Column-Prior Tiling Strategy}
\label{sec:tiling}
\begin{table}[t]  
	\small
	\caption{Notation for tiling strategy and dataflow.}  
	\vspace{1em}
	\label{lab:param}
	\begin{center}  
		\begin{tabular}{c|c}
			\hline  
			Variables & Descriptions \\ \hline  
			$W_T$ & width (column) of a tile of feature maps  \\ \hline  
			$H_T$ & height (row) of a tile of feature maps  \\  \hline
			%$N_w$ & number of input tiles to transfer \\ \hline
			$K_T$ & parallelism on output channel dimension \\ \hline
			$C_T$ & parallelism on the input channel dimension \\ \hline
			$N_k$ & number of tiles along the filter dimension \\ \hline
			$N_c$ & number of tiles along the channel dimension\\ 
			\hline  
		\end{tabular}  
	\end{center}  
\end{table}  

\begin{figure}
	\begin{center}
		\includegraphics[width=\columnwidth]{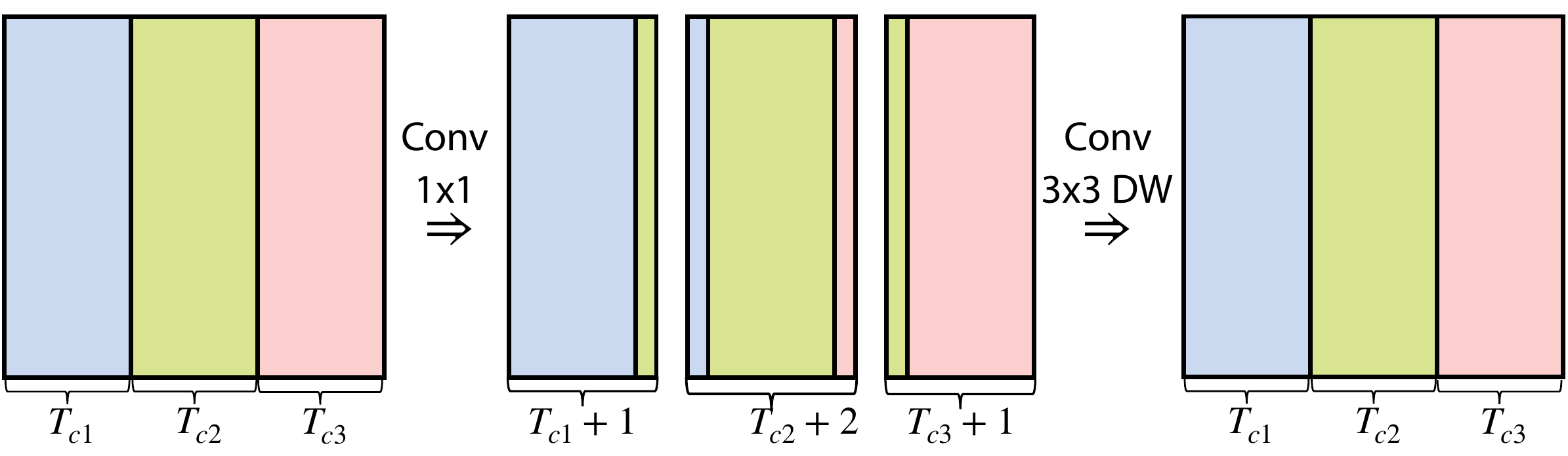}
	\end{center}
	\caption{A particular case of column-prior tiling strategy applied to 1$\times$1 and depthwise convolution (stride=1). The input feature maps are divided into three tiles and transferred from DDR to on-chip BRAMs sequentially. As shown in the middle of the figure, extra features columns from adjacent tiles are necessary.} 
	\label{tiling}
\end{figure}

%To map convolutional layers to the accelerator, tiling is necessary to accommodate the limited on-chip resources. 
Under the limited on-chip resources, tiling is necessary to map convolutional layers to the accelerator. 
We adopt a column-prior tiling strategy, as shown in Figure \ref{tiling}, which can reduce both latency and register consumption.  
We take a feature map with size 256$\times$256 as an example, which is expected to be divided into two parts to fit into the limited on-chip buffers. 
As for the row-prior manner, a tile with size 128$\times$256 is generated after 1$\times$1 convolution and can be sent to PE\_DW immediately for the processing of depthwise convolution. 
In this situation, at least $2\times256+3=515$ registers are required for applying line buffer convolution. 
However, if the feature maps are divided into the size of $256\times128$ in a column-prior manner, only $2\times128+3=259$ registers are needed. 
Thus register consumption can be approximately halved. 
Similarly, invalid cycles caused by filling registers are also reduced, which will also be beneficial to latency and efficiency.

\begin{algorithm}[t]
	\For{$h=0:H_T$}{
		\For{$w=0:W_T$}{
			\For{$n_k=0:N_k$}{
				\For{$n_c=0:N_c$}{
					\textbf{Parallel} 
					\For{$k=(n_k$-$1)K_t:n_kK_t$}{
						\textbf{Parallel} \For{$c=(n_c$-$1)C_T:n_cC_T$}{
							$P$ = $I[c][w][h]*W[k][c]$;\\
							$O[k][w][h]$ = $O[k][w][h]$ + $P$;\\
							// Partial sums keep stationary in PE until a valid output is obtained; \\
	}}}
	// Sent $O[(n_k$-$1)K_t:n_kK_t][w][h]$ to Out\_buf;\\
	}}}
	\caption{Output stationary dataflow for 1$\times$1 convolution.}
	\label{alg:output}
\end{algorithm}

\begin{algorithm}[t]
	\For{$n_k=0:N_k$}{
		\For{$n_c=0:N_c$}{
			// Fetch weights from weight buffer and keep weights stationary in PEs;\\
			\For{$h=0:H_T$}{
				\For{$w=0:W_T$}{
					\textbf{Parallel} \For{$k=(n_k$-$1)K_t:n_kK_t$}{
						\textbf{Parallel} \For{$c=(n_c$-$1)C_T:n_cC_T$}{
							$P$ = $I[c][w][h]*W[k][c]$;\\
							$O[k][w][h]$ = $O[k][w][h]$ + $P$;\\
							// Keep partial sums in Inter buffer; \\
		}}}}}
		// Sent $O[(n_k$-$1)K_t:n_kK_t][:][:]$ to out\_buf;\\
	}
	\caption{Weight stationary dataflow for 1$\times$1 convolution.}
	\label{alg:weight}
\end{algorithm}

Since the feature maps are divided into several tiles by column index,  overlapping between adjacent tiles are introduced. 
Suppose that we can obtain output tiles with five valid columns after 1$\times$1 and depthwise convolution (stride=1), the input tiles should contain seven valid values in each row. 
During the processing, a column of input features from the last tile is needed.

\subsection{Hybrid Dataflow}

Although column-prior tiling strategy is utilized for the efficiency of the accelerator, the on-chip buffer requirement and memory accesses depend heavily on the dataflow of computations \cite{chen2016eyeriss,chen2019eyeriss}.
The output stationary, as well as the weight stationary, is the most 
%common existing dataflow used in previous designs.  
commonly used dataflow in previous designs.
Algorithm \ref{alg:output} and \ref{alg:weight}  illustrate both dataflows, respectively, where the parameters are shown in Table \ref{lab:param}.

\begin{itemize}
	\item \textbf{Output stationary dataflow.} %Partial sums are held in PEs until we get the final results, while input feature maps and weights flow into PEs continuously.
	%And the results are either passed to PE\_DW or stored back to the IARAMs/OARAM.
	Input activations and weights are fed into the PE array continuously, and the partial sums are held in PEs until the final results are available. 
	These final results are either passed to PE\_DW for the following computation or stored in the IARAMs/OARAM.
	%Since each output is completed after all the weights in a filter have been calculated, higher bandwidth is required for weight transmission. 
	Since each output is completed after weights in a filter have been calculated, higher bandwidth is required for weight transmission.
	In addition, because of the implementation of weight buffer, there are more opportunities for weights to be reused.
	\item \textbf{Weight stationary dataflow.} Each PE holds part of weights for reuse until finishing the computation with input activations in the corresponding channels. And the partial sums generated in each PE are stored to the Inter RAM. Only if the kernel group is completed can the final results be sent to IARAMs/OARAM. In this way, weights can be reused as many as possible, but the accelerator requires additional storage, i.e., Inter RAM.
\end{itemize}

Although our accelerator is specialized for compact detection network, different convolutional layers (1$\times$1 convolution and depthwise convolution) still present heterogeneous property (e.g. width, height, and channel size). 
The dimensions of feature maps near to the input are relatively large. Thus these layers require more on-chip buffer to store the activations, while weights require less storage. 
In this case, there are more opportunities for weights to be reused, which is more suitable for output stationary dataflow.
However, in the deeper layers, weights become much more intensive in memory, because output stationary dataflow needs to fetch all the weights of a kernel to the PE to calculate each output. 
If the weight buffer can not accommodate those weights, weights are required to be fetched multiple times during the processing, leading to more energy consumption. 
In other words,  we need a larger weight buffer to reuse weights.

Therefore, we consider a hybrid dataflow that makes a balance between the weight reuse and weight buffer requirements to get the best performance and energy on the resource-limited computing platform. 
In most of the early layers, we adopt the output stationary dataflow. 
Thus, all the weights of a kernel group can be reused in weight buffer, and they are fetched from WRAM only once during the processing of a layer.
The case becomes different as the network goes deeper, and the weight stationary dataflow is adopted. So the weight buffer requirement can be significantly reduced with only a small Inter RAM overhead.

With the help of Co-Processor, our accelerator is flexible enough to support these two types of dataflow according to the size of kernels.

\section{Experiments} \label{sec5}

\begin{table}[t]
	\small
	\caption{Configurations of each type of PEs and the overall resource utilization on Ultra96 development board.}  
	\begin{center}  
		\begin{tabular}{c|c|c|c|c}
			\hline
			Parameters& $K_t$ & $C_T$&Precision(A/W/O)& Operations\\
			\hline
			PE\_33& 8 & 3 & 8/3/4 bits& Conv 3$\times$3 \\
			\hline 
			PE\_11&16&32& 4/3/4 bits& Conv 1$\times$1 \\
			\hline 
			PE\_DW & 16 & 16 & 4/3/4 bits & Conv 3$\times$3 DW \\
			\hline
			PE\_Head & 2 & 32 & 4/8/16 bits & Conv 1$\times$1 \\
			\hline
			
		\end{tabular}
	\end{center}  
	\noindent
	\begin{center}  
		\begin{tabular}{c|c|c|c}
			\hline  
			Resource&Available  &Used   &Utilization    \\ \hline  
			LUT     & 70560     &50485  &71.55\%          \\ \hline  
			FF      & 141120    &74174  &52.56\%          \\ \hline
			BRAM    & 216       &178.50 &82.86\%          \\ \hline
			DSP     & 360       &83     &23.06\%          \\ \hline
		\end{tabular}  
	\end{center}  
	\label{uti}
\end{table}  

We implement our solution on the Ultra96 development board with Xilinx Zynq UltraScale+ MPSoC. 
The accelerator runs at a frequency of 215 MHz with clock gating to each type of PE. 
Power measurement is obtained via a power monitor. 
We measured the power of approximate 6.9W on the Ultra96 when processing the detection task with  the image size of $512\times 512$. 
The configurations of each type of PE and the overall resource utilization are shown in Table \ref{uti}, in which we also list the supported precision of activations (A), weights (W), and outputs (O) respectively. 
%We can utilize less than 25\% of the total DSPs 
It shows that less than 25\% of the total on-chip DSPs are used on the FPGA since most of the multiplications are implemented as shift operations using LUTs. 
Most of the registers are used as weight buffer while BRAMs are mainly used for data buffer and the WRAM. 
With limited programmable resources on Ultra96 board, the whole system reaches an inference speed of 18 fps. 
Results are reported when the system is detecting objects from a video. 
Table \ref{spe} shows the specification of the entire system.

\begin{figure*}
	\begin{center}
		\includegraphics[width=1.7\columnwidth]{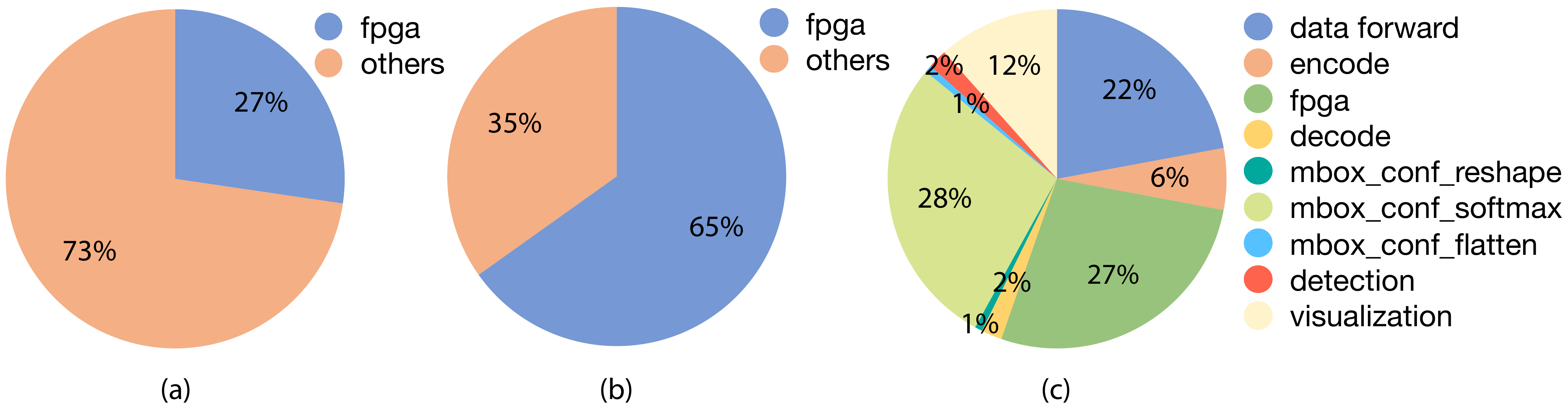}
	\end{center}
	\caption{(a) Time breakdown before pipeline. (b) Time breakdown after pipeline. (c) Detailed time breakdown of each layer before pipeline. Figure (a) and (b) demonstrate that with the heterogeneous pipeline, the overall latency is reduced, thus the proportion of FPGA layers becomes larger, dominates most of the latency. Figure (c) shows that data forward layer and mbox-conf-softmax layer are the most time-consuming layer and require more threads to process. } 
	\label{time}
\end{figure*}

%To complete the entire process of object detection, location and confidence information extracted from the accelerator requires post-processing which is processed on the ARM CPU. 
%Besides, data preprocessing is also performed by CPUs. 
%Since the ARM core on the Ultra96 board is designed for low-power applications, we need to do some work to speed up the processing of the data. 
%Though FPGA complete most of the calculation, 
Although FPGA undertakes most of the computations in detection algorithm, we find that pre-processing and post-processing on CPUs still account for most of the inference time, as shown in Figure \ref{time} (a). 
In order to overcome the bottleneck of CPU execution, %we adopt a pipelined manner with the help of multi-threads. 
we adopt a pipelined task management with multi-thread techniques.
In this way, the total latency is reduced, and FPGA layers dominate most of the inference time, as shown in Figure \ref{time} (b).

%Thread assignments are conducted empirically. Figure \ref{time} (c) presents the detailed time breakdown of each layer. As shown in the figure, the softmax layer is the most time-consuming among all the layers, while the decode layer and mbox-conf-reshape layer only accounts for 3\% of the latency. Therefore, we assigned a single core to these three layers. Data forward layer is also time-consuming and occupies a single core. Encode layer and FPGA layer are assigned to another core. The last remaining core is shared by other layers. The latency will vary greatly depending on the image because the number of objects can vary significantly with input changes. Therefore, time breakdown in Figure \ref{time} is obtained by averaging over a batch of images. 

Thread assignments are conducted empirically. 
Figure \ref{time}(c) presents the detailed time breakdown of each layer. 
The latency can vary greatly depending on the input image %because the number of objects can vary significantly with input changes. 
because the number of objects within an image varies significantly and thus influence the computational complexity in the post-processing phase.
Therefore, time breakdown in Figure \ref{time} is obtained by averaging over a batch of images. 
As shown in the figure, the softmax layer is the most time-consuming among all the layers, while the data forward layer and visualization layer account for 34\% of the latency. 
%And other layer take up negligible time.
Note that in a real-world application such as ADAS, the detection results are used as part of the control system, in which visualization may not be necessary. In this situation, the latency of CPUs can be further reduced, pushing the system frame rate towards the maximum.
\begin{table}[t]
	\small
	\caption{System specification.}  
	\vspace{1em}
	\begin{center}  
		\begin{tabular}{c|c}
			\hline  
			Device          & Ultra96 development board \\ \hline
			Network         & customized MobileNet-SSD         \\ \hline  
			Quantization    & activation 4b/weight 3b          \\ \hline
			Power           & 6.9Watt                   \\ \hline
			Frame rate      & 18 fps                    \\ \hline
			Accelerator frame rate & 27 fps             \\ \hline
			mAP on VOC 2012            & 66.4                      \\ \hline
		\end{tabular}  
	\end{center}  
	\label{spe}
\end{table}  

\begin{figure}[t]
	%\begin{center}
	\centering
	\includegraphics[width=0.9\columnwidth]{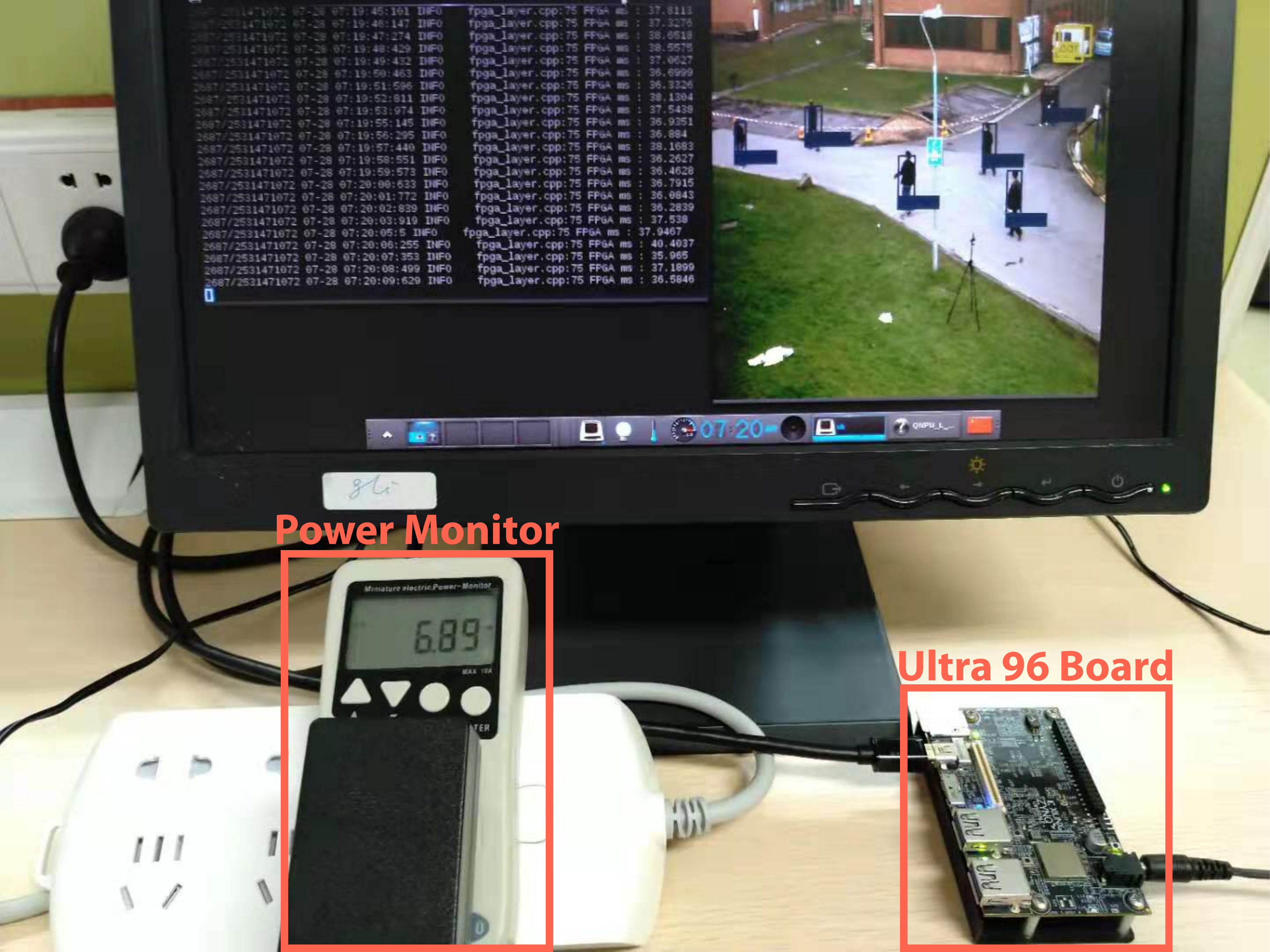}
	%\end{center}
	\caption{A demo of our proposed object detection system.}
	\label{object}
\end{figure}

%The entire system is practical to be used to process video streams. 
Figure \ref{object} shows a demo of our proposed object detection system. As we can see, the measured power is around 6.9W, and there are slight fluctuations as the detected image changes. Most of the targets are correctly detected (e.g. pedestrian, cars), frame rate for FPGA layers is around 25-30.

\begin{table}[h]
	\centering
	\caption{Comparison with other accelerators}
	\resizebox{\columnwidth}{!}{
		\begin{tabular}[h]{|c|c|c|c|c|}
			\hline
			& VGG\_ACC\cite{GoingDeeper} & Low-Bit\cite{DoReFa} & Synetgy\cite{ShiftAcc} & Ours \\ 
			\hline
			\textbf{Precision}&\multirow{2}{*}{16/16 bits}&\multirow{2}{*}{2/1 bits}&\multirow{2}{*}{4/1 bits}&\multirow{2}{*}{4/3 bits}\\
			\textbf{(A/W)} & & & & \\
			\hline
			\multirow{2}{*}{\textbf{Platform}}&Zynq&Zynq&Zynq &Zynq  \\
			&XC7Z045 &XC7Z020 &ZU3EG & ZU3EG\\
			\hline
			\textbf{Frequency}&\multirow{2}{*}{150}& \multirow{2}{*}{200}&\multirow{2}{*}{250}&\multirow{2}{*}{215}\\
			\textbf{(MHz)} & & & & \\
			\hline
			\textbf{Network} & VGG-16 & DoReFa & ShuffleNetV2 & MobileNet\\
			\hline 
			\textbf{Classification}& \multirow{2}{*}{64.64\%} & \multirow{2}{*}{46.10\%} &\multirow{2}{*}{68.47\%} & \multirow{2}{*}{68.1\%}\\
			\textbf{Top\_1 Acc} & &  & & \\
			\hline 
%			\textbf{Top\_5 Acc} & 86.66\% & 73.1\% & 88.22\% & \\ 
%			\hline 
			\textbf{Performance}&\multirow{2}{*}{136.97}&\multirow{2}{*}{410.22}&\multirow{2}{*}{47.09$\sim$418}&\multirow{2}{*}{202.76}\\
			\textbf{(GOPs)} & & & & \\
			\hline
		\end{tabular}	
	}
	\label{tab:result}
\end{table}

As shown in Table \ref{tab:result}, we also compare our accelerator against previous works. Since the previous works are mainly designed for image classification, we also evaluate the performance of our customized MobileNet on ImageNet classification task. Compared with VGG\_ACC, which is implemented with 16-bits integers, our design can achieve better performance and accuracy even on a smaller FPGA. Low-Bit is implemented with lower bits, which leads to severe accuracy degradation. Synetgy uses shift operations to replace the spatial convolutions. It can achieve high accuracy with lower bits, i.e., 4-bits activations and 1-bit weights. However, our accelerator can achieve more stable performance with comparable accuracy.

\section{Conclusion} \label{sec6}
In this paper, we present a system-level solution for object detection on the heterogeneous embedded system. We quantize the compact detection network to low bits, which allows us to replace multiplications with efficient shift operations. A dedicated CNN accelerator is implemented to carry out convolution computation. In order to support the arbitrary size of input feature maps under limited resources, we adopt a column-prior tiling strategy to map the convolutional layer to the accelerator. Compared to row-prior tiling strategy, it can reduce both register consumption and latency. According to the heterogeneous properties of different layers, we provide a hybrid dataflow, with which we can flexibly reuse the partial sums or filter weights. Multi-thread is also exploited to accelerate the pre-processing and post-processing. We believe that such an efficient and low energy system can play a role in IoT applications.

\section*{Acknowledgment}
This work was supported by the Strategic Priority Research Program of Chinese Academy of Sciences (Grant No. XDB32050200) and National Natural Science Foundation of China (Grant No.61972396, 61906193).

{\small
	\bibliographystyle{ieee}
	\bibliography{egbib}
}

\end{document}